\documentclass[runningheads]{llncs}
\usepackage{algorithm}
\usepackage{algpseudocode}
\usepackage{cite}
\usepackage{amsmath,amssymb,amsfonts}

\usepackage{graphicx}
\usepackage{textcomp}
\usepackage{xcolor}
\usepackage{graphicx}
\usepackage{latexsym}

\usepackage{microtype}
\usepackage{tikz}
\usepackage{mathrsfs}
\usepackage[hidelinks,pdfa]{hyperref}
\usepackage[T1]{fontenc}
\usepackage{booktabs}
\usepackage{float}
\usepackage{multicol}
\usepackage{booktabs}

 \newcommand{\squishlist}{
	\begin{list}{$\bullet$}
		{ \setlength{\itemsep}{0pt}
			\setlength{\parsep}{3pt}
			\setlength{\topsep}{3pt}
			\setlength{\partopsep}{0pt}
			\setlength{\leftmargin}{1.5em}
			\setlength{\labelwidth}{1em}
			\setlength{\labelsep}{0.5em}}}
	\newcommand{\squishlisttwo}{
		\begin{list}{$\bullet$}
			{ \setlength{\itemsep}{0pt}
				\setlength{\parsep}{0pt}
				\setlength{\topsep}{0pt}
				\setlength{\partopsep}{0pt}
				\setlength{\leftmargin}{2em}
				\setlength{\labelwidth}{1.5em}
				\setlength{\labelsep}{0.5em}}}
		\newcommand{\squishend}{
	\end{list}  }

\begin{document}
\title{ComplicaCode: Enhancing Disease Complication Detection in Electronic Health Records through ICD Path Generation}
%
%
\author{Xiaofan Zhou\inst{1}}
\authorrunning{Xiaofan}
%
\institute{
Worcester Polytechnic Institute, Worcester, USA\\
\email{xzhou5@wpi.edu}}
\maketitle              
\begin{abstract}
The target of Electronic Health Record (EHR) coding is to find the diagnostic codes according to the EHRs. In previous research, researchers have preferred to do multi-classification on the EHR coding task; most of them encode the EHR first and then process it to get the probability of each code based on the EHR representation. However, the question of complicating diseases is neglected among all these methods. In this paper, we propose a novel EHR coding framework, which is the first attempt at detecting complicating diseases, called ComplicaCode. This method refers to the idea of adversarial learning; a Path Generator and a Path Discriminator are designed to more efficiently finish the task of EHR coding. We propose a copy module to detect complicating diseases; by the proposed copy module and the adversarial learning strategy, we identify complicating diseases efficiently. Extensive experiments show that our method achieves a 57.30\% ratio of complicating diseases in predictions, and achieves the state-of-the-art performance among cnn-based baselines, it also surpasses transformer methods in the complication detection task, demonstrating the effectiveness of our proposed model. According to the ablation study, the proposed copy mechanism plays a crucial role in detecting complicating diseases.

\end{abstract}
\section{Introduction}
In the contemporary era of healthcare, the abundance of biomedical data has increasingly become crucial, making coding for Electronic Health Records (EHR) a significant task \cite{miotto2018deep}. EHR coding is a classification task that generates International Classification of Diseases (ICD) codes for EHRs, which plays a crucial role in various areas, e.g., search, data mining, billing, epidemiology, and disease detection for healthcare manufacturers \cite{choi2016doctor}. Deep learning methods for EHR coding are popular due to their higher accuracy and ability to handle large volumes of data efficiently within a short time frame.

Although many works on EHR coding have achieved success, EHR coding still remains a challenging task today, and multi-label classification has been the main trend in recent years' research on the EHR coding task \cite{kavuluru2015empirical,shi2017towards}. Among these methods, deep learning techniques, e.g., multilayer perceptron (MLP) \cite{taud2018multilayer}, and sequential models like long short-term memory and recurrent neural networks \cite{medsker2001recurrent,graves2012long}, generate probabilities for each code. Codes with probabilities higher than a threshold are then used as labels for the EHRs. However, these methods face several shortcomings.
Firstly, codes exist in a very high-dimensional space; in ICD-9-CM, there are more than 18,000 codes, and these codes have a severely unbalanced distribution.
Secondly, existing methods often neglect the fact that some codes may form a group of complications; their only target is to classify the codes directly. Complication often arises in medical diagnosis, indicating that one illness can lead to another. For example, people with diabetes mellitus may also have diabetic retinopathy and cataracts caused by diabetes.
Moreover, determining the appropriate threshold in these methods is challenging.
A second family of EHR coding methods generates a path where the code in this path is treated as the classification result for the EHR. This approach, called RPGNet, reduces the candidate space of ICD codes and uses each node as a prediction result for EHRs \cite{wang2020coding}. However, this method still cannot effectively address the problem of identifying complicating diseases. Recently, as transformer and pretrained methods appears, these methods are also applied into EHR coding area. In PLM-ICD\cite{huang2022plm},  they proposed to use pretrained language models to extract the representation of EHR text, however, this kind of work only use the pretrained language models to extract better representation of EHR but neglects other obstacles in the EHR coding area like complication detection.

To address these questions in existing methods, we treat the coding generation task as a path generation task. We generate labels for the EHR one by one, and each point on the path is seen as a label for the text. Then we propose to use a copy module to detect complications in the EHR. The proposed methods detect complications by involving the complication information among the ICD codes. It contains two major parts: one is the generate-mode, which is designed to generate new codes based on EHR and generated codes; another is the copy-mode, designed to increase the complication detection rate by considering complication disease information. Thus, the module can help the model detect complications.

In this paper, we concentrate on EHR coding methods aimed at identifying groups of complicating diseases. Our proposed model contains two parts, a path generator and a path discriminator. In the path generator, our new method generates ICD codes path using Recurrent Neural Networks (RNNs). When generating ICD codes path, we calculate the probability of each code based on EHR text and previous generated codes, then we choose the code with largest probability for every timestamp on the path. We employ the odds ratio to evaluate whether two nodes might be complications. Subsequently, to detect these complicating diseases, we use existing ICD codes to generate new ICD codes based on their complication information. This formulation allows for the discovery of ICD codes that are based on complication information. Furthermore, each output in the RNN can be considered as a prediction result, enabling the RNN method to be used for multi-label prediction without the need to determine a threshold for labels. And we use path discriminator to evaluate the accuracy of the path generated by path generator. The path discriminator and path generator are trained alternately to be optimized by each other.

The result of our experiments shows our model outperforms other cnn-based EHR coding methods, and the complication detection rate outperforms all of the baselines including pretrained EHR coding methods. These finding shows our model successfully deal with the above difficulties.

In summary, the major contributions of our paper are as follows:
\squishlist
\item We proposed a novel EHR coding methods, which focusing on detecting complication disease by. As far as we know, this is the first attempt to consider complication disease into EHR coding area.

\item We proposed a copy module for complication detection, which not only consider the complication information, but also consider the global information. It successfully achieves the goal to detect complication.

\item We conducts extensive experiments on MIMIC-III dataset. The result shows our model's superior, especially on complication detection ability. It also achieve SOTA performance among CNN encoder baseline.
\squishend 
\section{RELATED WORK}
\subsection{Automatic EHRs coding}
Deep learning has been proven efficient in various fields. The field of automatic EHR coding has recently seen a surge in publications, encompassing a range of approaches and techniques. Several recent studies address this question using sequential models. Li and Yu leverage a residual structure and a series of convolutional layers to extract more efficient representations for EHRs \cite{li2020icd}. \cite{choi2016doctor} and \cite{baumel2017multi} used recurrent neural networks (RNNs) to extract the representation of EHRs.

Recently, the RAC model \cite{kim2021read} proposed a Read, Attend, and Code (RAC) framework that can predict ICD codes more efficiently, greatly improving the task. In JointLAAT \cite{vu2020label}, a developed label-aware attention mechanism is leveraged in a bidirectional LSTM. Hypercore \cite{cao2020hypercore} considers the hierarchical structure as a hyperbolic space and leverages a graph-based neural network to capture co-occurrences of codes. ISD \cite{zhou2021automatic} proposes a self-distillation learning mechanism to alleviate the long-tail code distribution and uses shared representation extraction for codes of different frequencies. EffectiveCAN \cite{liu2021effective} developed a network with a squeeze-and-excitation (SE) mechanism connected by residual connections and extracted representations from all encoder layers for label attention. They also introduced focal loss to address data imbalance issues. However, these recent models aim to find the relationship between EHR representations and code representations. 

As the surge of large language model and pretrained model, many works begin to use pretrained model or large language model to extract the representation of EHR text, and predict their labels based on the representation. PLM-ICD \cite{huang2022plm} use pretrained bert to code extract the representation of EHR coding. EPEN \cite{teng2024few} proposes to use meta-network to implement few shot task in EHR coding. TreeMAN \cite{liu2023treeman} fuse tabular features and textual features into multimodal representations by enhancing the text representations with tree-based features via the attention mechanism.

However, these works neglect the importance of complication detection in EHR coding. In contrast, our work emphasizes the relationship between the codes, applying a copy module to identify complicating diseases.

\subsection{Adversarial learning in healthcare}
Generative Adversarial Networks (GANs) have demonstrated impressive results in generative tasks, especially in computer vision. A GAN comprises a discriminator and a generator, where the generator creates outputs to deceive the discriminator, and the discriminator is designed to classify which outputs are from the generator and which are from the ground truth. In the healthcare domain, various GAN models have excelled in Electronic Health Record (EHR) tasks. RGAN and RCGAN \cite{esteban2017real} generate realistic time-series data, while medGAN \cite{choi2017generating} and mtGAN \cite{guan2018generation} produce convincing EHRs. Furthermore, \cite{yu2019rare} deals with rare disease detection using semi-supervised GANs. LDP-GAN \cite{gwon2024ldp}  prevent the leakage of training data by protecting the model from malicious attacks using local differential privacy(LDP). 

Distinct from previous research, our approach employs seqGAN concepts \cite{yu2017seqgan}, integrating GAN with LSTM to generate multi-label outputs and calculate rewards for every generated path. The discriminator evaluates the legitimacy of the path, using the outcomes as rewards for each generated path, thereby guiding our model's learning process.

\section{METHOD}
The aim of this study is to formulate a series of diagnostic codes, denoted as \( Y = [c^1, c^2, \ldots, c^j, \ldots] \), which correspond to Electronic Health Records (EHRs). Each \( c^i \) in this series signifies an International Classification of Diseases (ICD) code and where i means the position of code in the path, and \( c^i \in C \), where $C$ represents code dictionary.
For the sake of simplification, we treat an EHR, denoted as \( x \), as unstructured text. In mathematical terms, \( x = [w_1, w_2, \ldots, w_i, \ldots] \), where \( w_i \in W\), W is the words token dictionary. 
The methods we propose to achieve this objective will be elaborated in the subsequent sections of this paper.


\begin{figure*}[!htb]
  \centering
  \includegraphics[width=\textwidth]{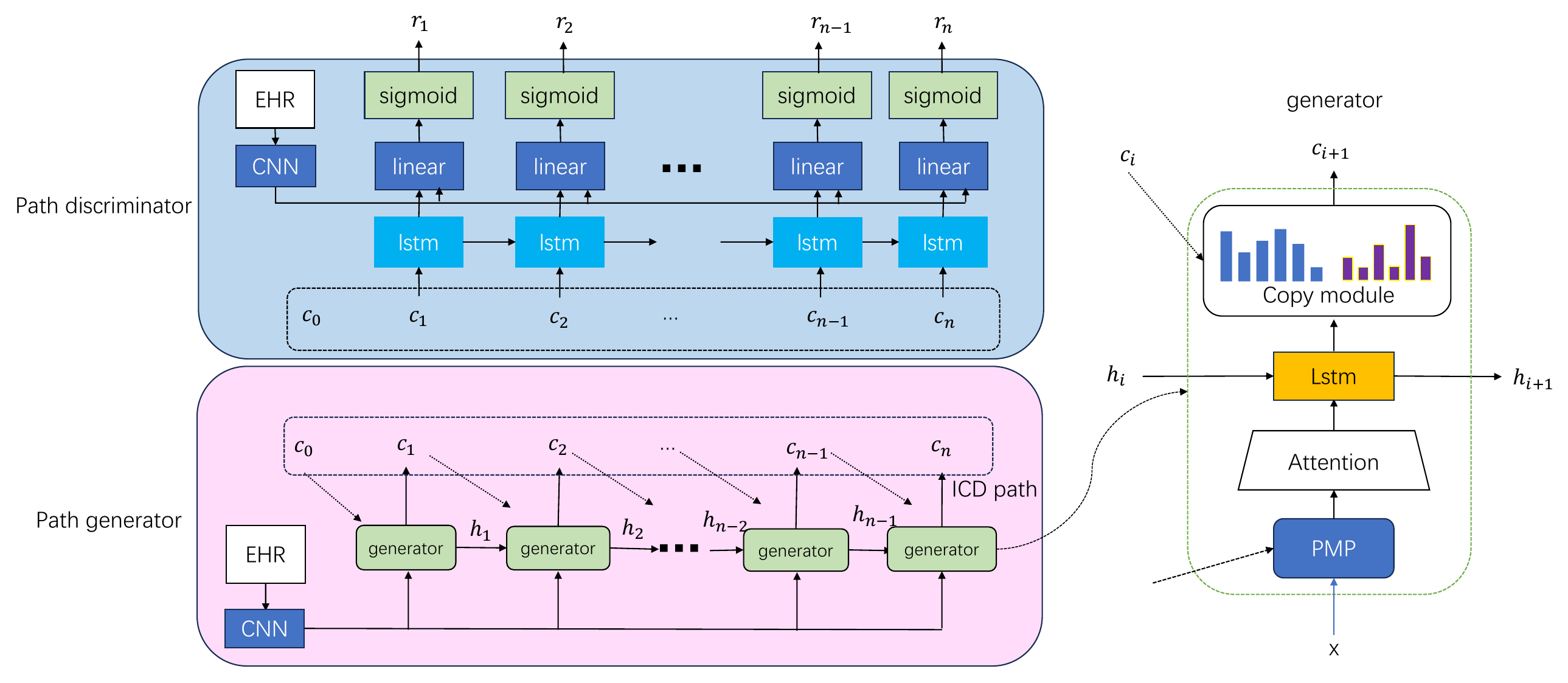}
  \caption{Overview of ComplicaCode}
  \label{fig:your_image_label}
\end{figure*}

\subsection{Overview}
In ComplicaCode, there are two crucial parts, the first part is a path generator\( G_\theta \), the path generator is used to generate ICD path, the ICD path will be seen as the classification result for the given EHR text $x$, in the path generator\( G_\theta \), firstly, a EHR encoder extract feature representation for EHR text $x$ using convolutional neraul network and a token embedding matrix $I^{n*d}$ where $n$ represents the count of the token while $d$ represents  embedding matrix's dimension. Then, we propose a EHR-ICD fusion module, to combine the information of previous generated code, EHR embedding. To consider the long-short term relation, we leverage LSTM module to generate the probability of every code at each timestamp and choose the code with largest probability as the code at each timestamp, finally, a copy module is used to help with finding the complication disease based on the preceding code $c_{t-1}$. We use the probability given by LSTM and copy module as the classification result. The length of LSTM is a preset hyper-parameter. Overall, the goal of our proposed model is to predict the ICD codes from discharge summaries in electronic health record.

To classify whether the code path generated by the path generator \( G_\theta \) is a feasible code path, we apply the idea of GAN to design a path discriminator \( D_\phi \), which output a probability score \( r_t \), \( r_t \) means the probability of truth of the generated path. The input of \( D_\phi \) is the embedding of EHR \( x \), and the ICD path is generated by path generator\( G_\theta \). By the way, we split the ICD path generated by path generator\( G_\theta \) into $t$ parts, where $t$ is the valid codes number in the path, so that we can enlarge the sample number fed into the \( D_\phi \). This path discriminator helps to better train the path generator, and updated in a certain round.

\subsection{Long Short Term Memory Networks (LSTM)}
Recurrent Neural Network (RNN) architecture is the main part of the path generator, specialized for processing sequential data, in this paper, RNN is used to process the generated code path. Unlike traditional feedforward neural networks, RNNs are designed with interconnected hidden states that allow them to capture dynamic temporal patterns within sequences. This structure makes RNNs particularly adept at identifying long-range temporal dependencies. Compared to other RNNs, Long Short-Term Memory (LSTM) networks is special. Each neuron in LSTMs contains three gates: input\( i \),forget\( f \), and output \( o \). These gating mechanisms bestow LSTMs with three distinct advantages: firstly, they excel in learning long-term sequence dependencies; secondly, they facilitate easier optimization, as the gates effectively route the input signal through recurrent hidden states \( r(t) \) without causing undesired alterations in the output; and thirdly, they mitigate the well-known exploding and vanishing gradient encountered during the training of conventional RNNs. The max length of this LSTM is a predefined hyperparameter in this paper.

The mathematical representation of the LSTM is given by:
\begin{equation}
\begin{aligned}
f_t &= \sigma(W_f \cdot [h_{t-1}, c_t] + b_f) \\
i_t &= \sigma(W_i \cdot [h_{t-1}, c_t] + b_i) \\
\tilde{C}_t &= relu\left(W_C \cdot [h_{t-1}, c_t] + b_C\right) \\
C_t &= f_t \times C_{t-1} + i_t \times \tilde{C}_t \\
o_t &= \sigma(W_o \cdot [h_{t-1}, c_t] + b_o) \\
h_t &= o_t \times \tanh(C_t)
\end{aligned}
\end{equation}
$\sigma$ denotes sigmoid function, $\cdot$ represents element-wise product, and the $W_*$ matrix is weight matrix, $b_*$ is bias vector, $h_{t-1}$ represents hidden representation before. In this paper, rectified linear units (ReLU) is used for activation function.

\subsection{Path generator \texorpdfstring{$G_\theta$}{go}}
The path generator is constituted by four parts: we first get the representation $x$ by multichannel CNN encoder from raw text, then we use EHR-ICD fusion module to consider the connection between the codes in the path and get $v$ as representation, after put the $v$ into lstm, then we use the former code in the path $c$ and $o$ into the copy module to get the final codes which can help with finding the complications. We will introduce this progress step by step.
\subsubsection{EHR-ICD fusion module}
We use the EHR-ICD fusion module to process the hidden state $h_t$ obtained from LSTM by considering the relationship between EHR $x$ and the generated Codes path, the relationship between the current code and all other codes. Next, we will talk about how EHR-ICD fusion module is used to address the question.

\paragraph{relationship between EHR and the generated Codes path} In this step, we fuse the representation of EHR $x$ and the representation of generated code path, and we represent the result in this step as $o_t$, which is shown below:
\begin{equation}
\begin{aligned}
\label{eq2}
o_t = tanh(W_o(u_t))
\end{aligned}
\end{equation}
where $W_o$ is weight matrix and $u_t$ is calculated by a series of operation on the representation of $x$ and $o_t$, such operation is shown below:
\begin{equation}
\begin{aligned}
\label{eq3}
u_t = x\oplus c_t \oplus (x\odot c_t)\oplus (x+c_t)\oplus (x-c_t)\oplus (c_t-x)
\end{aligned}
\end{equation}
The $\oplus$ represents concatenation operation, $\odot$ represents product.
\paragraph{Former-to-later code relationship} This part aims to capture the relationship of previous ICD codes $c_{t-1}$ and subsequent ICD codes $c_{t}$. To find the how generated code and all codes in relation, we use the $u_t$ generated by considering the relationship between an EHR and the generated Codes path as input, we use the representation of all codes to do the task before:
\begin{equation}
\begin{aligned}
\label{eq4}
v_t = o_t \cdot c_t^c
\end{aligned}
\end{equation}
$c_t^c$ is the representation of each code, $\cdot$ is the element-wise product operation.

\subsubsection{Copy module for complication detection}
To detect complication disease, we propose a novel module called copy module. The copy module consists of two major parts, one is generate-mode, which is designed for generating new code based on global information including EHR text and existing generated codes, another one is copy-mode, this one is used to generate new code based the prior code with its complication information. The final result of this module is the combination of these two modes. We will introduce whole process of copy mudule in this parts.

First of all, a code vocabulary $\mathcal{V_{c}} = \{v_1, ..., v_N\}$ is designed for code $c$ as its potential complication disease and use UNK for any code have low probability to be complication. As $\mathcal{C}$ is the whole code dictionary, considering code of $\mathcal{C}$ let this module consider non-complication diseases. 
Then Given the decoder lstm state $h_{t-1}$ at time $t-1$, we generate the 'mixture' probability of code $c_t$ as follows:
\begin{equation}
\begin{aligned}
    & p\left(c_{t} \mid \mathbf{h}_{t}, c_{t-1}\right) \\
    & =p\left(c_{t}, c \mid \mathbf{h}_{t}, c_{t-1}, \mathcal{V_{c_{t-1}}}\right)+p\left(c_{t}, g \mid \mathbf{h}_{t}, c_{t-1},\mathcal{C}\right)
\end{aligned}
\end{equation}
where $g$ represents the generate-mode, and $c$ denotes the copy mode. The probabilities for these two modes are given respectively by:
\begin{equation}
\begin{gathered}
\label{eq8}
    p\left(c_{t}, c \mid \cdot\right)=\left\{\begin{array}{cc}
    \frac{1}{Z} e^{\psi_{c}\left(h_{t}\right)}, & y_{t} \in \mathcal{V} \\
    0, & y_{t} \in \mathcal{X} \cap \bar{V} \\
    \end{array}\right.
\end{gathered}
\end{equation}

\begin{equation}
    \begin{gathered}
    \label{eq9}
        p\left(c_{t}, g \mid \cdot\right)=
        \frac{1}{Z} \sum_{j: c_{j}=y_{t}} e^{\psi_{g}\left(h_{j}\right)}\\
    \end{gathered}
\end{equation}

$\psi_{g}\left(.\right)$ and $\psi_{c}\left(.\right)$ are rating methods, and we treat $Z$ as the normalization term $Z=\sum_{v \in \mathcal{V} \cup{\mathrm{UNK}}} e^{\psi_{g}(v)}+\sum_{x \in X} e^{\psi_{c}(x)}$. Caused by this normalization term, above two modes essentially compete through a softmax function. The score of each mode is calculated:

\paragraph{Generate-Mode:}
As our final objective is to predict the label for each EHR, we can not neglect the influence brought by EHR text and should not only consider the influence of previous code. In the generate-mode, we predict the next code in the path based on the global information of EHR text and previous codes in the path. The rating function used is same as in the generic RNN encoder-decoder \cite{bahdanau2014neural},
\begin{equation}
    \label{eq10}
    \psi_{g}\left(y_{t}=v_{i}\right)=\mathbf{v}_{i}^{\top} \mathbf{W}_{o} \mathbf{h}_{t}, \quad c_{i} \in \mathcal{C}
\end{equation}
where $\mathbf{W}_{o} \in \mathbb{R}^{(N+1) \times d_{s}}$ is a weight matrix and $v_i$ is one-hot representation of $c_i$.
\paragraph{Copy-Mode:}In the copy-mode, we design this mode to consider the influence of complication disease by introduce the complication information into path generation process. The rating for copy module is calculated as
\begin{equation}
    \begin{gathered}
    \label{eq11}
\psi_{c}\left(c_{t}=x_{j}\right)=\sigma\left(\mathbf{h}_{j}^{\top} \mathbf{W}_{c}\right) \mathbf{h}_{t}, \quad x_{j} \in \mathcal{V}
    \end{gathered}
\end{equation}

In this equation, $\mathbf{W}{c} \in \mathbb{R}^{d{h} \times d_{s}}$ is the weight matrix, and $\sigma$ is activation function. Empirically, we discovered that non-linearity activation tanh performs better than other activation function, and we utilize the one-hot complicating disease vector for $h_t$.

\subsection{EHR encoder}
In this section, we extract the representation $x$ of each EHR for the sake of later processing by multi-channel CNN\cite{chen2015convolutional}. The first step in this encoder is embedding the tokens in the EHR into a embedding matrix $X\in R^{n \times d}$ $n$ is the total number of tokens among all EHR and $d$ is the dimension of the embedding matrix. In the convolutional layers, we apply $c_i$ different filters to each kernel size $k_i, 1 \leq i \leq m$. Each filter $f_{ij}$, $1 \leq j \leq c_i$, is a matrix $\in \mathbb{R}^{k_i \times d}$, and it is used for $k_i$ words in text to produce a feature map $\mathbf{F}_{ij} \in \mathbb{R}^{(n - k_i + 1) \times 1}$:

\begin{equation}
\begin{array}{ll}
\mathbf{F}_{ij}[p] = \text{ReLU}(\mathbf{f}_{ij} \cdot \mathbf{X}_{p:p+k_i-1} + b_{ij}),
\end{array}
\end{equation}

where $\cdot$ denotes the dot product, $\mathbf{X}_{p:p+k_i-1}$ is the $p$-th to $(p+k_i-1)$-th row of $\mathbf{X}$, $b_{ij}$ is a bias term, and ReLU is nonlinear activation function.

Next, we utilize max-pooling to each feature map $\mathbf{F}_{ij}$ to obtain a scalar $s_{ij}$:

\begin{equation}
\begin{array}{ll}
s_{ij} = \max_{p} \mathbf{F}_{ij}[p].
\end{array}
\end{equation}

Finally, we concatenate all the $s_{ij}$ as the final representation of the text:

\begin{equation}
\begin{array}{ll}
\mathbf{x} = [s_{11}, s_{12}, \ldots, s_{m c_m}].
\end{array}
\end{equation}
\subsection{Overall objective function}
Given the above result, the final objective function for our model is calculate by following equation:
\begin{equation}
\label{eq:loss}
\begin{array}{ll}
J(\theta) = \mathbb{E}_{a \sim \pi(a|h,x;\theta)} \left( \sum_{t} R_{(h_t,x) c^t} \right)\\

= \sum_{t} \pi(c^t | h_t, x; \theta) R_{(h_t,x) c^t}
\end{array}
\end{equation}
The $R_{(h_t,x) c^t}$ is reward given by discriminator in eq \ref{eq:re}, $c^t$ is the code at timestamp $t$ on the generated code path, $h_t$ is the hidden representation at timestamp $t$, $x$ is the representation of EHR text, $\pi(c^t | h_t, x; \theta)$ is the probability of $c^t$ is a label of $x$ calculated by generator $G_\theta$.
\subsection{Orderless Recurrent Models}
As code for each EHR is orderless, we need to find a way to compare our generated code which is in order, we adopt the concept from \cite{yazici2020orderless} to align them with the network's predictions before calculating the loss. We use the predicted label alignment (PLA) to achieve this goal.

The core idea behind PLA is when predicted label is true, we prefer to keep it. Which leads to the following optimization problem:

\begin{equation}
\begin{array}{ll}
\label{eq12}
\mathcal{L}= \qquad \min\limits_{T} & \operatorname{tr}(T \log (P)) \\
\text { subject to } & T_{t j} \in\{0,1\}, \sum_{j} T_{t j}=1 \\
& T_{t j}=1 \text { if } \hat{l}_{t} \in L \text { and } j=\hat{c}_{t} \\
& \sum_{t} T_{t j}=1 \ \forall j \in L \\
& \sum_{t} T_{t j}=0 \ \forall j \notin L
\end{array}
\end{equation}
\( \hat{c}_t \) represents the model's predicted label at time step \( t \). Initially, we secure the elements in matrix \( T \) corresponding to predictions that are verified by the ground truth set \( L \). Subsequently, we employ the Hungarian algorithm to allocate remaining labels. This approach offers greater alignment with the labels that the LSTM model actually predicts.

\subsection{Path Discriminator \texorpdfstring{$D_\phi$}{dq}}

For the sake of training the path generator in a more efficient way, we refer from the idea of GAN to design a path discriminator module $D_\phi$, $D_\phi$ is designed to generate a reward $r_t$ for every pairs of EHR representation $x$ and path representation generated by a path encoder, ICD code path $y_t=(c_1,c_2,...,c_k,...,c_t)$ by path generator we generate the path representation by lstm encoder:
\begin{equation}
\begin{array}{ll}
h_k &= LSTM(h_{k-1}, c_k),
\end{array}
\end{equation}
$h_{k-1}$ is the hidden state at timestamp $k-1$; $c_k$ is the $k$-th ICD code representation.
Then the calculation of $r_t$ is shown below:
\begin{equation}
\label{eq:re}
\begin{array}{ll}
r_{t} &= R_{\left(s_{t}, x\right), a_{t}} \\
&= p_{D}\left(y_{t}, \boldsymbol{x}\right) \\
&= \sigma\left(\boldsymbol{W}_{r}\left(\boldsymbol{h}_{t} \oplus \boldsymbol{x}\right)\right),
\end{array}
\end{equation}
$\oplus$ represents concatenation, $\boldsymbol{W}_{r}$ represents weight of a linear layer. The path $y_t$ is obtained by recurrently applying an LSTM to the ICD code path $y_t=(c_1,c_2,...,c_k,...,c_t)$, and $\sigma$ represents the sigmoid function.

We train the path discriminator using adversarial training strategy. Here, the paths generated by the path generator are treated as negative samples, the ground truth paths are considered as positive samples. We optimize a cross-entropy function to improve the effectiveness of $D_\phi$:

\begin{equation}
\label{eq:dis}
\begin{array}{ll}
L_{D} &=-\sum_{\left(y_{t}, x\right) \in Q^{+}} \log p_{D}\left(y_{t}, x\right) \\
&-\sum_{\left(y_{t}, x\right) \in Q^{-}} \log \left(1-p_{D}\left(y_{t}, x\right)\right),
\end{array}
\end{equation}
$Q^{+}$ and $Q^{-}$ represent a pair of positive and negative samples, respectively; $p_D\left(y_t,x\right)$ is the probability of sample ($y_t,x$) belonging to a positive sample. 

\section{Experimental Design}
\subsection{Research Questions}
We plan to solve the following research problems:

(RQ1)How well ComplicaCode preform on finding the complicating disease. 

(RQ2) What are the effects of different components? Where do the improvements of ComplicaCode come from?
\subsection{Dataset}
We use MIMIC-III as the dataset for comparison, which is a large EHR dataset. Similar with previous studies\cite{mullenbach-etal-2018-explainable}, we only focus on discharge summaries in EHR, which use one document to summarize information of patients. Therefore, We clean discharge summaries by removing noisy information, such as doctors’ information and hospital information. We also tokenize the remaining text and use stopword removal and stemming to preprocess the EHR. To simplify the problem, we use a top-50 label setting to preprocess the dataset.

For the top-50 label setting, we only use the most frequent 50 codes from the dataset and filter out instances that do not contain at least one of these codes. This results in a subset of the dataset that is more focused and relevant to our task, as we only focus on improving the complication disease detection ability by our model.

The split ratio of the train, test, and validation datasets from the resulting dataset is 4:1:1, with each set containing approximately 80\%, 10\%, and 10\% of the instances, respectively. This allows us to evaluate the performance of our models on unseen data while ensuring the training set is large enough for effective learning.

\begin{table*}[htbp]
	\centering
	\caption{Results(\%) on dataset, bold face indicates the best result.All result has p-value<0.01.}
    \resizebox{\linewidth}{!}{
	\begin{tabular}{lcccccccccc}
		\toprule
		& & &\multicolumn{2}{c}{Precision} & \multicolumn{2}{c}{Recall} & \multicolumn{2}{c}{F1} & \multicolumn{2}{c}{AUC} \\
		Method & Jaccard & Complication & Micro & Macro & Micro & Macro & Micro & Macro & Micro & Macro \\
		\midrule
        CAML(CNN) & 27.68 & 24.57 & 58.59 & 38.35 & 40.25 & 33.50 & 47.63 & 33.59 & 68.31 & 66.35\\
        RPGNet(CNN) & 33.34 & 36.31 & 41.20 & 33.59 & 56.81 & 44.51 & 48.35 & 35.40 & 75.14 & 70.50 \\
        
        ComplicaCode(CNN) &  \underline{34.59} & \textbf{57.30} & 54.67 & 34.74 & 43.59 & 32.01 & \underline{48.45} & 31.31 & 69.58 & 65.36 \\
        \midrule
        PLM-ICD(Bert) & 54.87 & 42.29 & 68.26 & 64.24 & 71.25 & 66.74 & 69.72 & 65.46 & 92.64 & 90.14\\
        \bottomrule
	\end{tabular}%
    }
	\label{tabl}%
\end{table*}%

\begin{table*}[htbp]
	\centering
	\caption{Ablation study}
    \resizebox{\linewidth}{!}{
	\begin{tabular}{lcccccccccc}
		\toprule
		& & &\multicolumn{2}{c}{Precision} & \multicolumn{2}{c}{Recall} & \multicolumn{2}{c}{F1} & \multicolumn{2}{c}{AUC} \\
		Method & Jaccard & Complication & Micro & Macro & Micro & Macro & Micro & Macro & Micro & Macro \\
		\midrule
        ComplicaCode & 34.59 & \textbf{57.30} & 54.67 & 34.74 & 43.59 & 32.01 & 48.45 & 31.31 & 69.58 & 65.36 \\
        ComplicaCode(no-copy) & 35.05 & 51.86 & 54.64 & 35.03 & 45.39 & 31.93 & 49.52 & 31.28 & 70.39 & 65.15 \\
        ComplicaCode(no-ARL) & 34.09 & 55.80 & 54.57 & 34.10 & 44.73 & 31.79 & 49.08 & 30.94 & 70.08 & 65.10 \\
		\bottomrule
	\end{tabular}%
    }
	\label{tabl2}%
\end{table*}%
\subsection{Baselines}
To demonstrate the performance and functionality of ComplicaCode, several methods are re-implemented for comparison. Specifically, we compare ComplicaCode with following methods:
\paragraph{CAML\cite{mullenbach2018explainable}}
CAML leverages Text-CNN\cite{chen2015convolutional} to generate representations for each Electronic Health Record (EHR). It then employs attention module to learn salient features for coding. This tailored representation forms the basis for binary classification tasks.

\paragraph{RPGNet\cite{choi2017generating}}
RPGNet approaches EHR coding by employing a dual structure consisting of a Path Generator (PG) and a Path Discriminator (PD). They propose PMP module to encode the path. In addition, the PD component calculates a reward value for every generated path. The network is trained using adverserial learning strategy.

\paragraph{PLM-ICD\cite{huang2022plm}}
PLM-ICD utilizes pretrained Language Models to obtain context-rich representations. It use transformer methods the encode the EHR content. It currently stands as the state-of-the-art method for ICD classification.

\subsection{Metric}
In our study, we utilize a comprehensive set of evaluation metrics, frequently employed in related work~\cite{mullenbach2018explainable}. Our evaluation framework incorporates both micro- and macro-averaged metrics. The micro-averaged approach treats each instance as a separate prediction during computation, whereas the macro-averaged method averages metrics for each individual label, making it especially relevant for assessing performance on rare labels.
More specifically, we examine four primary metrics: Precision, Recall, F1 Score, and the Area Under the Curve (AUC)~\cite{goutte2005probabilistic}. The F1 Score is the harmonic mean of Precision and Recall, and the AUC provides a summary of the model's performance across various thresholds.
Additionally, we employ the Jaccard Similarity Coefficient as an overlap measure between the sets of predicted and actual ICD codes~\cite{niwattanakul2013using}. This coefficient is defined as 
\[ Jaccard = \frac{1}{m} \sum_{i=1}^{m} \frac{|Y_i \cap \hat{Y}_i|}{|Y_i \cup \hat{Y}_i|}, \]
\( m \) denotes the number of code in the dataset, while \( Y_i \) is the set of predicted codes, and \( \hat{Y}_i \) represents the set of actual codes.
Furthermore, we gauge the models' effectiveness in identifying complications by evaluating the proportion of complications in each disease combination present in the predictions. The formula for computing the complication ratio is as follows: 
\begin{equation}
\begin{array}{ll}
complication = \frac{\sum_{1<=i<j<=n} I(c_i,c_j)}{\binom{2}{n}}
\end{array}
\end{equation}
where $I(c_i,c_j)$ represent whether code $i$ and code $j$ is a pair of complication, $n$ represent the number of codes in the prediction.
\subsection{Implementation details}
We use 300 as the dimension for the representation of EHR and path, and the dimension of the token and code embedding is chosen from as \{64, 100, 128\}(Eq. 3 and Eq. 4). For the EHR encoder, the filter size of convolutional layers is 100, while the kernel size of these layers are 3, 4, 5, respectively. while the the drop ratio of the dropout is 0.5 (Eq. 11). For EHR-ICD fusion module, we set the hidden size of LSTM gate as 300. In training stage, we use 200 as the max train iteration, and the batch size is chosen from \{16, 32, 64\}. We train path generator and path discriminator in every batch alternately. All model parameters are randomly initialized. Adam optimizer\cite{kingma2017adam} is employed to train the objective function, learning rate chosen from \{0.0001, 0.001\} and default momentum parameters of $\beta_1$ = 0.9 and $\beta_2$ = 0.999. We train other baselines based on the hyperparameter provided in their paper. All codes are implemented in PyTorch and we train them on an NVIDIA T4 Tensor Core GPU.

\section{Experimental Result}
\subsection{RQ1.1}

We assessed the performance of various models, including Caml, RPGnet, ComplicaCode, and PLM-ICD. We evaluated these models using the top-50 label setting and trained the first three models with the same preprocessing method. Additionally, we incorporated a pretrained PLM-ICD (BERT) model in our analysis. The results presented in Table 1 lead to the following observations:

(1)Among the compared models, ComplicaCode achieved the highest complication ratio. Notably, ComplicaCode's complication ratio was 15.01\% higher than PLM-ICD, the state-of-the-art method for EHR coding. This indicates that ComplicaCode effectively identifies complications associated with other codes. This may be attributed to ComplicaCode's exploration of ICDs through a sequence of paths, as well as its utilization of information from previous codes to generate new ones.

(2)PLM-ICD is powered by BERT, a significantly more advanced approach for extracting text representations, utilizing a pretrained model that benefits from training on an larger dataset. This superiority underscores the correlation between a high-quality model and its capability in detecting complex diseases. Moreover, the rate of complication detection serves as a crucial metric for assessing a model's performance. While our model may lag behind PLM-ICD in terms of Jaccard, it excels in complication detection, a testament to both the efficacy of our model and the promise of our proposed module.

\subsection{Ablation study(RQ1.2)}

We conducted an ablation study to evaluate the effects of modules in ComplicaCode, shown in Table 2. Followings are ablation of ComplicaCode that we considered: (1) No copy represents ComplicaCode removed the copy module. We removed the copy module and used only the RNN structure model to generate codes. (2) No ARL represents ComplicaCode removed adversarial learning strategy. We removed the adversarial learning process to train our model. From the results, we obtained the following finding:

First, the ability of ComplicaCode to identify complication diseases decreased dramatically after removing the copy module. Specifically, the complication ratio in the result of ComplicaCode(no-copy) dropped by 5.44\% respectively. The result shows that copy module plays a significant role in our method to identify complicating diseases. The copy module extracts information related to complicating diseases based on the former ICD in the path, which helps to generate complicating diseases.

Second, the ability of ComplicaCode to identify complicating diseases is still working better than PLM-ICD when copy module is not used. The complication ratio of PLM-ICD is 42.29\% and that of ComplicaCode(no-copy) is 51.86\%, which indicates that the RNN structure is also helpful in identifying complicating diseases, as the ICD on timestamp $t+1$ is generated based on the ICD on timestamp $t$.

Finally, the adversarial reinforcement learning (ARL) does not significantly improve the performance of ComplicaCode, but it slightly improves its ability to identify complicating diseases. This is due to adversarial reinforcement learning helps ComplicaCode generate codes paths similar to the ground truth paths.
\section{CONCLUSION AND FUTURE WORK}
In this paper, we propose an innovative approach for Electronic Health Record (EHR) coding, specifically targeting the identification of complicating diseases. We frame this as a path generation task and introduce the ComplicaCode. ComplicaCode consists of a EHR-ICD fusion module designed to fuse the representation of EHRs and code path. An adversarial learning framework is used for training stage. Our experimental results on the MIMIC-III dataset reveal that ComplicaCode surpasses existing methods in detecting complicating diseases, and its performance excels when using the CNN encoder method. The performance improvement is attributed to the implementation of the copy module and the adversarial reinforcement learning mechanisms.

\bibliographystyle{abbrv}
\bibliography{paper}
\end{document}